\documentclass[conference]{IEEEtran}
\IEEEoverridecommandlockouts
\usepackage{cite}
\usepackage{amsmath,amssymb,amsfonts}
\usepackage{algorithmic}
\usepackage{graphicx}
\usepackage{textcomp}
\usepackage[table]{xcolor}
\usepackage{pifont}
\usepackage{tikz}
\usepackage{subcaption} 
\usepackage{graphicx}   

\usepackage{booktabs}
\def\BibTeX{{\rm B\kern-.05em{\sc i\kern-.025em b}\kern-.08em
    T\kern-.1667em\lower.7ex\hbox{E}\kern-.125emX}}
\begin{document}

\title{V-CORE: Temporally Consistent Video Understanding for Video-LLM}

\author{
\IEEEauthorblockN{
Zhengjian Kang$^{1}$,
Qi Chen$^{2}$,
Rui Liu$^{3}$,
Kangtong Mo$^{4}$,
Xingyu Zhang$^{5}$,
Xiaoyu Deng$^{6}$,
Ye Zhang$^{7}$\thanks{Corresponding author: Ye Zhang (yez12@pitt.edu).}
}
\IEEEauthorblockA{$^{1}$ New York University, USA}
\IEEEauthorblockA{$^{2}$ University of California, Irvine, USA}
\IEEEauthorblockA{$^{3}$ Illinois Institute of Technology, USA}
\IEEEauthorblockA{$^{4}$ University of Illinois Urbana-Champaign, USA}
\IEEEauthorblockA{$^{5}$ George Washington University, USA}
\IEEEauthorblockA{$^{6}$ Fordham University, USA}
\IEEEauthorblockA{$^{7}$ University of Pittsburgh, USA}
\IEEEauthorblockA{\small
Emails: zk299@nyu.edu, qic7@uci.edu, liuruiabc1@gmail.com, mokangtong@gmail.com,\\
xingyu\_zhang@gwmail.gwu.edu, xdeng24@fordham.edu, yez12@pitt.edu
}
}

\maketitle

\begin{abstract}
Recent Video Large Language Models (Video-LLMs) have shown strong multimodal reasoning capabilities, yet remain challenged by video understanding tasks that require consistent temporal ordering and causal coherence. Many parameter-efficient Video-LLMs rely on unconstrained bidirectional projectors to model inter-frame interactions, which can blur temporal ordering by allowing later frames to influence earlier representations, without explicit architectural mechanisms to respect the directional nature of video reasoning. To address this limitation, we propose V-CORE, a parameter-efficient framework that introduces explicit temporal ordering constraints for video understanding. V-CORE consists of two key components: (1) Learnable Spatial Aggregation (LSA), which adaptively selects salient spatial tokens to reduce redundancy, and (2) a Causality-Aware Temporal Projector (CATP), which enforces structured unidirectional information flow via block-causal attention and a terminal dynamic summary token acting as a causal sink. This design preserves intra-frame spatial interactions while ensuring that temporal information is aggregated in a strictly ordered manner. With 4-bit QLoRA and a frozen LLM backbone, V-CORE can be trained efficiently on a single consumer GPU. Experiments show that V-CORE achieves strong performance on the challenging NExT-QA benchmark, reaching 61.2\% accuracy, and remains competitive across MSVD-QA, MSRVTT-QA, and TGIF-QA, with gains concentrated in temporal and causal reasoning subcategories (+3.5\% and +5.2\% respectively), directly validating the importance of explicit temporal ordering constraints.
\end{abstract}



\section{Introduction}

Video understanding is inherently a temporally structured reasoning problem. Unlike static images, the semantic meaning of a video emerges from ordered event transitions and their causal dependencies over time. With the rapid progress of Large Language Models (LLMs) \cite{achiam2023gpt4, touvron2023llama2}, recent research has shifted from coarse activity recognition toward more complex video question answering (VQA) tasks that require explaining \textit{why} and \textit{how} events unfold \cite{liu2023llava, li2023blip2}. To enable efficient multimodal reasoning, many parameter-efficient Video-LLMs \cite{maaz2024videochatgpt, lin2023videollava, zhang2023videollama} adopt a two-stage pipeline: a frozen visual encoder extracts frame-level features, which are then mapped into the LLM token space via a lightweight projection module. This design allows effective video-language alignment without modifying the heavy LLM backbone.

\begin{figure}[t]
    \centering
    \includegraphics[width=\linewidth,height=4.5cm]{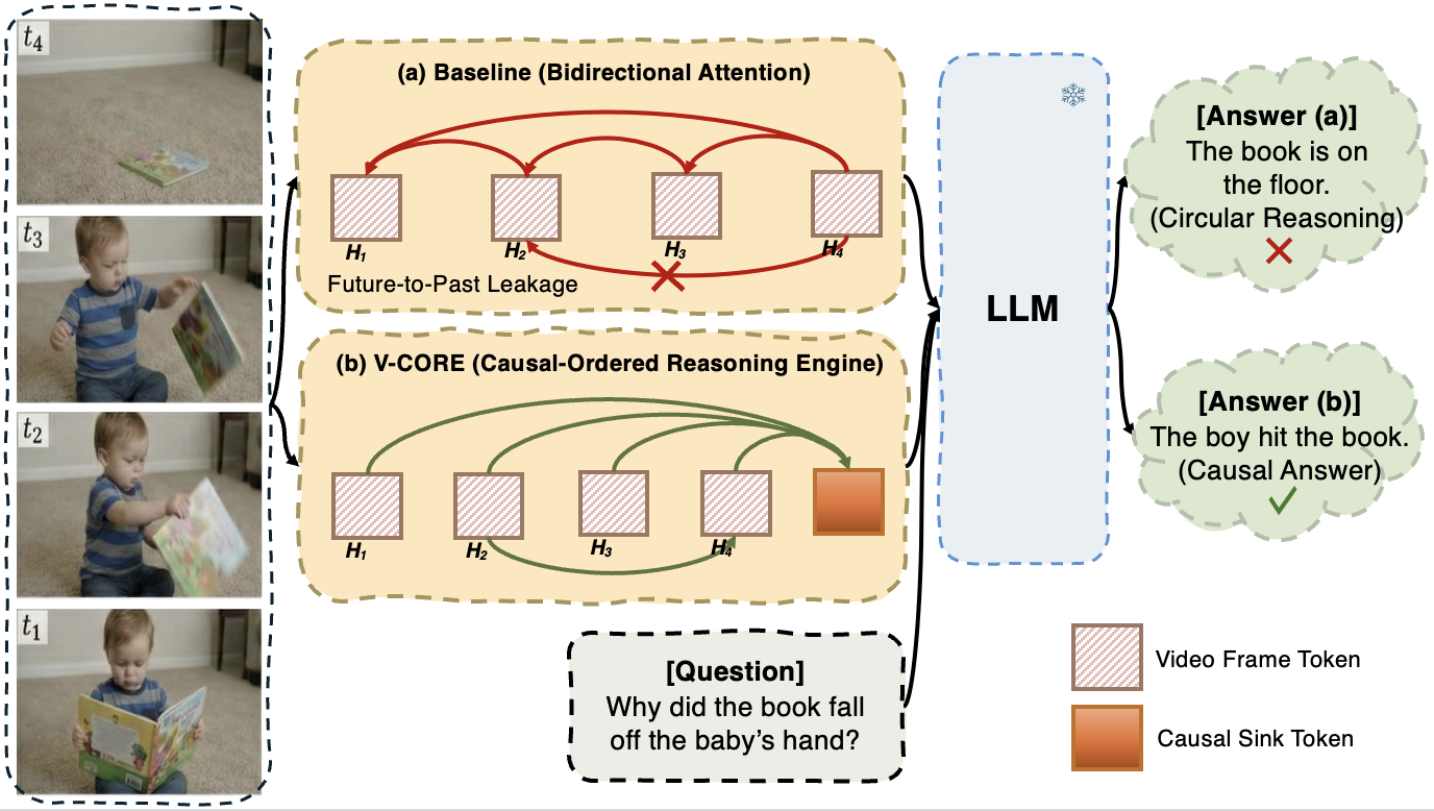} 
    \caption{\textbf{Baseline vs. V-CORE}. 
    \textbf{Baseline}: Unconstrained bidirectional attention may mix information across temporal directions. 
    \textbf{V-CORE}: Explicit unidirectional aggregation via a terminal causal sink enforces ordered temporal reasoning.}
    \label{fig:teaser}
\end{figure}

Despite their effectiveness, existing parameter-efficient Video-LLMs often lack explicit architectural mechanisms to model temporal ordering in a structured manner. To manage temporal dependencies under strict computational constraints, most approaches rely on heuristic pooling strategies or unconstrained bidirectional transformers to aggregate inter-frame information. 
While such designs are flexible and expressive, they may introduce \emph{temporal leakage}, where information from later frames influences representations of earlier events, thereby weakening ordered temporal reasoning. As illustrated in Fig.~\ref{fig:teaser}(a), this effect is particularly undesirable in benchmarks like NExT-QA~\cite{xiao2021next}, where reasoning about event order and causality is central to the task.

To address this limitation, as shown in \textbf{Fig.~\ref{fig:teaser}(b)}, we propose \textbf{V-CORE} (\textbf{V}ideo \textbf{C}ausal-\textbf{O}rdered \textbf{RE}asoning), a parameter-efficient framework that introduces explicit temporal ordering constraints into the video-to-LLM projection stage. 
The core of V-CORE is the \emph{Causality-Aware Temporal Projector} (CATP), which employs a block-causal attention mechanism to preserve intra-frame spatial interactions while enforcing strict inter-frame temporal ordering. Unlike standard causal masks applied to flattened token sequences, our block-causal design explicitly respects the hierarchical spatio-temporal structure of videos. 
In addition, we introduce a terminal dynamic summary token that functions as a causal sink, aggregating information only from past frames to form a coherent global representation without future-context interference. 
Complementing this design, our \emph{Learnable Spatial Aggregation} (LSA) module adaptively selects salient spatial tokens per frame, reducing redundancy and allowing the model to allocate its limited token budget to critical temporal transitions.

To ensure practical applicability, we adopt a 4-bit QLoRA \cite{dettmers2023qlora} training paradigm with LoRA adapters \cite{hu2021lora}. By freezing both the visual encoder and the LLM backbone and optimizing only lightweight modules, V-CORE enables end-to-end training on a single consumer GPU. Experimental results demonstrate the effectiveness of explicit temporal ordering. 
On the challenging NExT-QA benchmark \cite{xiao2021next}, V-CORE achieves 61.2\% accuracy, with the most notable improvements concentrated in the \emph{Temporal} and \emph{Causal} sub-categories, while maintaining competitive performance on descriptive questions.

Our main contributions are summarized as follows:
\begin{itemize}
    \item We highlight the lack of explicit temporal ordering constraints in existing Video-LLM projection modules and study its impact on temporal and causal reasoning.
    \item We propose the Causality-Aware Temporal Projector (CATP), which introduces block-causal attention and a terminal summary token to enforce structured unidirectional temporal aggregation.
    \item We develop Learnable Spatial Aggregation (LSA) to reduce spatial redundancy, enabling efficient long-form temporal modeling under a compact token budget.
    \item Extensive experiments show that V-CORE improves temporal and causal reasoning on NExT-QA while remaining efficient and competitive across multiple VideoQA benchmarks.
\end{itemize}

\section{Related Work}

\subsection{Video-Language Models}
Video-Language Models (Video-LLMs) extend the instruction-following capabilities of LLMs~\cite{achiam2023gpt4, touvron2023llama2,xing2025openemma} to the video domain by coupling a frozen vision encoder with a lightweight projection module.
Early systems such as Video-ChatGPT~\cite{maaz2024videochatgpt} and Video-LLaMA~\cite{zhang2023videollama} established this paradigm using linear or shallow projectors to map visual features into the LLM token space.
Subsequent works, including Video-LLaVA~\cite{lin2023videollava}, further improved multimodal alignment through instruction tuning and better visual-language pretraining.
More recent models such as LLaVA-NeXT~\cite{zhang2024llavanext} and Qwen2-VL~\cite{wang2024qwen2vl} scale video understanding by increasing visual granularity through dynamic resolution and advanced token compression strategies.
While these approaches achieve strong performance, temporal modeling is primarily handled implicitly within the LLM’s self-attention layers.
As a result, the projection stage is often agnostic to temporal directionality, treating frame-level features as unordered or symmetrically related tokens.

Complementary to video-centric modeling, fine-grained multimodal understanding of complex visual content has been studied in prior work~\cite{deng2024covis,wang2022internvideo}, highlighting the importance of structured visual representations for downstream reasoning.
Beyond accuracy, recent studies have also examined reliability aspects of multimodal models, including robustness under visual adversarial perturbations~\cite{cui2023robustnesslargemultimodalmodels,Li_2026_WACV} and stability under task-incremental learning settings where forgetting can occur~\cite{li2025mcl}.
In contrast, our work focuses on parameter-efficient video QA and introduces explicit architectural constraints at the projection stage to preserve temporal ordering, rather than relying on implicit temporal reasoning to emerge within the LLM.

\subsection{Spatio-Temporal Modeling and Token Compression}
Modeling temporal structure is central to video understanding, as it captures motion, event ordering, and causal dependencies.
Vision Transformers such as TimeSformer~\cite{bertasius2021timesformer} and ViViT~\cite{arnab2021vivit} explicitly model spatio-temporal interactions, but their computational cost limits their direct adoption as connectors for large language models. Structured attention and relation modeling~\cite{kang2025lpdetr,yu2024spatial,kang2026paq} have been explored in vision transformers to improve efficiency and representation quality. 
To maintain efficiency, many Video-LLMs compress frame-level features into a compact set of tokens before projection.
Recent methods such as ST-LLM~\cite{liu2024stllm} and LLaMA-VID~\cite{li2024llamavid} reduce redundancy through token pruning or aggregation, but typically rely on global or bidirectional attention mechanisms within the temporal projector. As highlighted by recent benchmarks such as Video-MME~\cite{fu2024videomme}, the absence of explicit directional constraints can hinder long-range temporal reasoning, particularly in tasks requiring event ordering. Unlike prior approaches that primarily focus on reducing token count, our method jointly addresses \emph{what} information to preserve and \emph{how} it is temporally structured. Specifically, Learnable Spatial Aggregation selects salient spatial tokens per frame, while the Causality-Aware Temporal Projector enforces a unidirectional temporal flow consistent with the arrow of time, culminating in a terminal causal sink for global aggregation.

\subsection{Parameter-Efficient Fine-Tuning (PEFT)}
Parameter-Efficient Fine-Tuning (PEFT) has become the de facto standard for adapting large backbones. LoRA~\cite{hu2021lora} and QLoRA~\cite{dettmers2023qlora} enable fine-tuning 7B+ models on consumer-grade hardware by injecting low-rank trainable matrices. While early Video-LLMs primarily trained the multimodal projector, recent methods have increasingly applied LoRA to the LLM layers to enable deeper modality alignment.
Our method builds upon the QLoRA paradigm but fundamentally revisits the architecture of the temporal connector. Instead of a frame-wise linear layer that treats frames as independent entities, we employ a causality-aware temporal transformer featuring a terminal causal sink to explicitly enforce unidirectional temporal information flow. This design mitigates temporal leakage that can arise from unconstrained bidirectional attention in temporally ordered video representations. This architectural refinement ensures that the visual context provided to the LLM is both temporally ordered and causally consistent.

\section{Method}

\begin{figure*}[!t]
    \centering
    \includegraphics[width=0.9\textwidth,height=0.3\textheight]{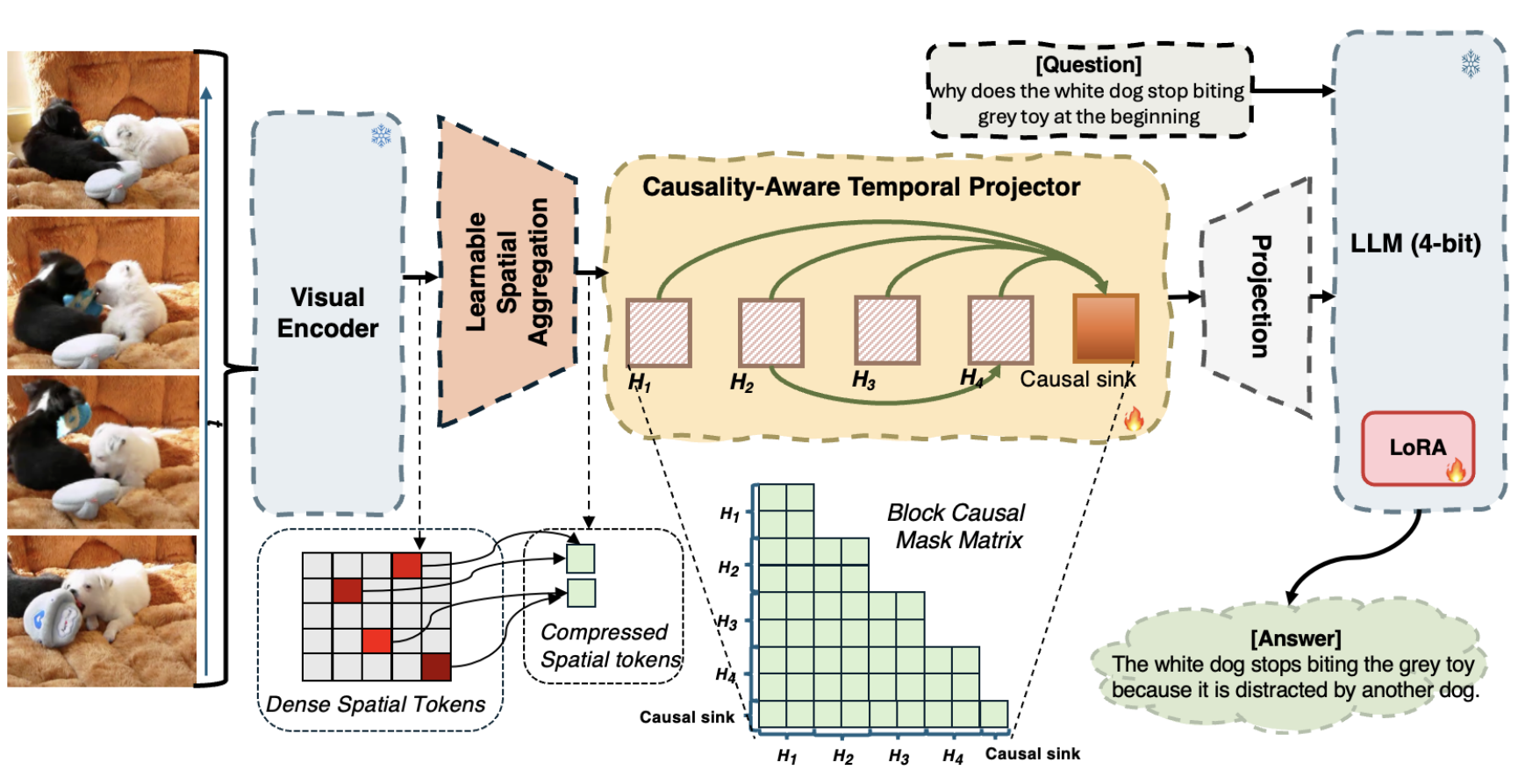}
    \caption{\textbf{Overall architecture of V-CORE.} Video frames are spatially compressed into salient tokens $H_t$ via LSA and temporally modeled by CATP under a block-causal constraint. The Causal Sink aggregates the sequence for reasoning in the LLM with LoRA.}
    \label{fig:arch}
\end{figure*}

\subsection{Problem Formulation}
Given a video $V$ and a textual instruction $X$, our goal is to generate a video-grounded response $Y$. To capture the full temporal extent of the video, we represent it as a sequence of $T$ frames $V = \{f_t\}_{t=1}^T$. Specifically, we employ a segment-based uniform sampling strategy, where the video is divided into $T$ equal segments, and one frame is sampled from each segment. 
This ensures that $V$ captures a temporally representative sequence of events.
We model the conditional distribution as $P(Y \mid V, X; \theta)$, using a frozen vision encoder (CLIP ViT-L/14) and a frozen Vicuna-7B v1.5 as the large language model (LLM) backbone. All learnable capacity is dedicated to the spatio-temporal alignment and causality-aware temporal reasoning modules.

\subsection{Architecture Overview}
As illustrated in Fig.~\ref{fig:arch}, we propose \textbf{V-CORE} (\textbf{V}ideo \textbf{C}ausal-\textbf{O}rdered \textbf{RE}asoning), a temporal-dominant framework designed for robust event-ordering and temporal-consistency. Our architecture consists of three core components:
(1) a \emph{Learnable Spatial Aggregation} (LSA) module that preserves $K$ tokens per frame to capture fine-grained spatial details, 
2) a \emph{Terminal Dynamic Summary} (Causal Sink) appended to encapsulate holistic context, and
(3) a \emph{Causality-Aware Temporal Projector} (CATP) that enforces an explicit unidirectional information flow.

\subsection{Learnable Spatial Aggregation (LSA)}
To handle spatial redundancy, we propose the \emph{Learnable Spatial Aggregation} (LSA) module. Each sampled frame $f_t$ is encoded into $N=256$ patch embeddings $F_t \in \mathbb{R}^{N \times d_v}$. Instead of the average pooling, LSA employs a set of $K$ learnable query tokens $\mathbf{Q} \in \mathbb{R}^{K \times d_v}$ to adaptively select salient spatial regions. 
Specifically, the aggregation is formulated as a cross-attention mechanism. For each frame $f_t$, the queries are derived from $\mathbf{Q}$, while the keys and values are projected from the spatial patches $F_t$:
$Q_s = \mathbf{Q} W^Q$, $K_s = F_t W^K$, $V_s = F_t W^V$, where $W^Q$, $W^K$, $W^V$ $\in \mathbb{R}^{d_v \times d_v}$ are learnable weight matrices. The aggregated spatial features $H_t$ are then computed as:
\begin{equation}
    H_t = \text{Softmax} \left( \frac{Q_s K_s^T}{\sqrt{d_k}} \right) V_s \in \mathbb{R}^{K \times d_v},
\end{equation}
where $d_k$ denotes the key dimension and is set equal to $d_v$. By retaining $K$ distinct tokens per frame ($K \ll N$), LSA preserves multi-faceted spatial semantics—such as object positions and salient foregrounds—while significantly reducing the input sequence length.

\subsection{Terminal Dynamic Summary}
\label{sec:summary}
To provide the LLM with a holistic understanding of the video narrative while adhering to temporal consistency, we move beyond static heuristics and introduce a Terminal Dynamic Summary mechanism. Specifically, we define a learnable query token $\mathbf{q}_{\text{sum}} \in \mathbb{R}^{d_v}$ acting as a \textit{causal sink}. 
Prior to the temporal projector, we append $\mathbf{q}_{\text{sum}}$ to the end of the spatio-temporal sequence $H_v = [H_1, \dots, H_T]$ derived from LSA module. The resulting augmented input sequence is formulated as:
\begin{equation}
    H_{\text{in}} = [{H}_1, \dots, {H}_T, \mathbf{q}_{\text{sum}}] \in \mathbb{R}^{(TK+1) \times d_v},
\end{equation}
where $T$ and $K$ denote the number of sampled frames and aggregated spatial tokens per frame, respectively. This positioning ensures that $\mathbf{q}_{\text{sum}}$ can adaptively aggregate information from the entire preceding temporal process. Due to the block-causal mask in CATP (Sec.~\ref{sec:catp}), this summary token is explicitly prohibited from influencing earlier representations, thereby preserving the temporal integrity of the frame-level features.

\subsection{Causality-Aware Temporal Projector (CATP)}
\label{sec:catp}

CATP does not assume ground-truth causality, but enforces a directional inductive bias that aligns the projector with the temporal structure of videos. The CATP module is the core engine for modeling cross-frame dependencies while respecting the arrow of time at the architectural level. It implements a multi-layered causality-aware architecture to bridge the gap between fragmented spatial tokens and coherent causal narratives.

\noindent\textbf{Spatio-Temporal Positional Encoding.} 
To preserve the structural hierarchy of the $(TK+1)$ tokens, we employ a joint spatio-temporal positional embedding $\mathbf{P} \in \mathbb{R}^{(TK+1) \times d_v}$. For a token at index $i$, the embedding is defined as:
\begin{equation}
    \mathbf{P}_i = \text{E}_{\text{temp}}(\lfloor i/K \rfloor) + \text{E}_{\text{spat}}\!\left(i \bmod K\right),
\end{equation}
where $\text{E}_{\text{temp}}$ and $\text{E}_{\text{spat}}$ represent learnable temporal and spatial encodings. For the terminal $\mathbf{q}_{\text{sum}}$, we assign a dedicated temporal index $T$ and a summary-specific spatial embedding distinct from all frame-level tokens to avoid indexing ambiguity.

\noindent\textbf{Causal Attention Mechanism.} 
The sequence $\mathbf{X}_0 = \mathbf{H}_{\text{in}} + \mathbf{P}$ passes through $L$ layers of causal transformer blocks. To enforce the temporal causal constraint, we apply a \textbf{block-causal mask} $\mathbf{M} \in \{0, -\infty\}^{(TK+1) \times (TK+1)}$ during the self-attention operation:
\begin{equation}
    \mathbf{M}_{ij} = 
    \begin{cases} 
    0, & \text{if } \lfloor i/K \rfloor \ge \lfloor j/K \rfloor \\
    -\infty, & \text{if } \lfloor i/K \rfloor < \lfloor j/K \rfloor
    \end{cases}.
\end{equation}

This formulation allows intra-frame interactions while strictly prohibiting future-to-past inter-frame dependencies. As the terminal element, $\mathbf{q}_{\text{sum}}$ dynamically aggregates the culmination of the temporal process, effectively serving as a causal anchor for the entire sequence.

\noindent\textbf{Visual-Language Interface.} 
After $L$ layers of reasoning, the refined features $\mathbf{X}_L \in \mathbb{R}^{(TK+1) \times d_v}$ are projected into the LLM's semantic space:
\begin{equation}
    \mathbf{Z}_v = \mathbf{W}_{\text{proj}} \mathbf{X}_L \in \mathbb{R}^{(TK+1) \times d_{\text{llm}}},
\end{equation}
where $\mathbf{W}_{\text{proj}}$ maps the features to the LLM embedding space. 
The resulting representation $\mathbf{Z}_v$ provides the LLM with a dual-perspective context: the fine-grained tokens ground reasoning in specific event transitions, while the terminal anchor $\mathbf{h}_{\text{global}}$, corresponding to the final hidden state of the summary query $\mathbf{q}_{\text{sum}}$, offers a causally-consistent global narrative summary.

\subsection{Parameter-Efficient Adaptation}
We utilize QLoRA~\cite{dettmers2023qlora} to insert low-rank adapters into the attention layers of the frozen Vicuna-7B. This enables the LLM to adapt to the structured temporal tokens produced by CATP, rather than learning video representations from scratch. The model is optimized using the auto-regressive loss on general video-instruction datasets, enabling robust zero-shot generalization. This design ensures that performance gains primarily stem from the proposed spatio-temporal modeling modules rather than increased model capacity.

\section{Experiments}

\begin{table}[t]
\centering
\caption{Statistics of the Video QA benchmarks.}
\label{tab:datasets}
\resizebox{\linewidth}{!}{%
\begin{tabular}{l|c|c|c|c}
\toprule
\textbf{Dataset} & \textbf{\# Videos} & \textbf{\# QA Pairs} & \textbf{Avg. Duration} & \textbf{Temporal Focus} \\
\midrule
MSVD-QA & 1.9K & 50K & $\sim$10s & Objects \& Simple Actions \\
MSRVTT-QA & 10K & 243K  & 10--30s & Open-Domain Events \\
TGIF-QA & $\sim$100K GIFs & 165K & $\sim$5s & Fine-Grained Motion \\
NExT-QA & 5K & 50K & $>30$s & Causal \& Temporal Reasoning \\
\bottomrule
\end{tabular}}
\end{table}

\subsection{Datasets and Metrics}
We evaluate our V-CORE framework on four VideoQA benchmarks (Table~\ref{tab:datasets}). \textbf{MSVD-QA}~\cite{ye2017video} and \textbf{MSRVTT-QA}~\cite{xu2016msr} focus on short-range visual grounding. \textbf{TGIF-QA}~\cite{jang2017tgif} demands sensitivity to motion patterns (\textit{Action} and \textit{Transition} subsets). Our primary benchmark, \textbf{NExT-QA}~\cite{xiao2021next}, requires complex reasoning to explain event dependencies in long-form videos. 
Following Video-ChatGPT~\cite{maaz2024videochatgpt}, we report Accuracy (Acc) and a semantic score (0--5 scale) for open-ended questions. Since video-language models may generate responses with varying lengths and expressions, traditional word matching metrics are insufficient. We therefore adopt an LLM-assisted evaluation protocol, where GPT-3.5-turbo judges the correctness and semantic relevance of predicted answers given the question and ground-truth response.

\subsection{Implementation Details}
\noindent\textbf{Model Configuration.} 
We employ Vicuna-7B v1.5 and a frozen CLIP ViT-L/14 encoder. Videos are sampled at $T=16$ frames and resized to $224 \times 224$. 
The LSA module uses $K=16$ learnable queries, and CATP features $L=2$ transformer layers with heads $h=8$, and hidden dimension $d_v=1024$. 
A terminal global summary token is appended to the sequence of $TK$ frame-level tokens, forming a compact visual context with $TK+1$ tokens. 
The resulting visual features are projected to the LLM embedding space ($d_{\text{llm}}=4096$) via a linear projection layer before being fed into the LLM.

To ensure parameter efficiency, a 4-bit NormalFloat (NF4) QLoRA~\cite{dettmers2023qlora} is applied \textit{only} to the query ($W_q$) and value ($W_v$) projection layers of the LLM. We set the LoRA rank to $r=64$ with a scaling factor $\alpha=128$. 
This lightweight adaptation strategy ensures that performance gains primarily stem from the proposed spatio-temporal modeling modules.

\noindent\textbf{Training Strategy.}
Following a zero-shot protocol, V-CORE is instruction-tuned exclusively on the Video-ChatGPT-100K dataset for 3 epochs without exposure to any target VideoQA benchmark data. We use the AdamW optimizer with a peak learning rate of $5 \times 10^{-5}$, a cosine learning rate scheduler, and a warmup ratio of 0.03. 
The effective batch size of 16 is achieved via gradient accumulation on a single NVIDIA RTX 4090 (24GB). We enable gradient checkpointing to manage memory overhead. The entire training process completes within approximately 20 hours. During evaluation, the model is directly tested on all benchmarks without task-specific fine-tuning.

\begin{table*}[t]
\centering
\caption{Comparison with state-of-the-art methods on 4 zero-shot video QA benchmarks.}
\label{tab:main_results}
\resizebox{0.98\textwidth}{!}{
\begin{tabular}{l|c|c|cc|cc|cc|cc}
\toprule
{\textbf{Method}} & {\textbf{LLM}} & \textbf{Data} & 
\multicolumn{2}{c|}{\textbf{MSVD-QA}} & 
\multicolumn{2}{c|}{\textbf{MSRVTT-QA}} & 
\multicolumn{2}{c|}{\textbf{TGIF-QA}} & 
\multicolumn{2}{c}{\textbf{NExT-QA}} \\
& \textbf{Backbone} & \textbf{Scale} & Acc & Score & Acc & Score & Acc & Score & Acc & Score \\
\midrule
\textit{Efficient Baselines} & & & & & & & & \\
Video-LLaMA~\cite{zhang2023videollama} & LLaMA-7B & 150K & 51.6 & 2.5 & 29.6 & 1.8  & - & - & - & - \\
VISTA-LLaMA~\cite{ma2024vista} & LLaMA-7B & 100K & 65.3 & 3.6 & 60.5 & 3.3 & - & - & 60.7 & 3.4  \\
Video-ChatGPT~\cite{maaz2024videochatgpt} & Vicuna-7B & 100K & 64.9 & 3.3 & 49.3 & 2.8 & 51.4 & 3.0 & 54.6 & 3.2 \\
\rowcolor{gray!10} \textbf{V-CORE (Ours)} & \textbf{Vicuna-7B} & \textbf{100K} & \textbf{67.2} & \textbf{3.7} & \textbf{57.8} & \textbf{3.3} & \textbf{57.2} & \textbf{3.3} & \textbf{61.2} & \textbf{3.5} \\
\midrule
\textit{Large-Scale Reference} & & & & & & & & \\
VideoChat~\cite{li2023videochat} & Vicuna-7B & 10M & 56.3 & 2.8 & 45.0 & 2.5 & 34.4 & 2.3 & 56.2 & 3.2 \\
Video-LLaVA \cite{lin2023videollava} & Vicuna-7B & 700K & 70.4 & 3.9 & 58.3 & 3.5 & 70.0 & 4.0 & - & - \\
VideoChat2 \cite{li2024mvbench} & Vicuna-7B & 2M & 70.0 & 3.9 & 54.1 & 3.3 & 60.4 & - & 61.7 & - \\
\bottomrule
\end{tabular}
}
\end{table*}

\subsection{Main Results}
Table~\ref{tab:main_results} reports the zero-shot performance of V-CORE across four benchmarks, demonstrating a strong performance-efficiency trade-off, particularly for long-range temporal reasoning.

\noindent\textbf{Results on NExT-QA.}
Our V-CORE achieves 61.2\% accuracy on NExT-QA, surpassing VISTA-LLaMA (60.7\%) and approaching VideoChat2 (61.7\%) despite using substantially less training data (100K vs. 2M), validating the benefit of explicit temporal ordering under a zero-shot setting.

\noindent\textbf{Results on Other Benchmarks.}
Beyond long-form reasoning, V-CORE remains competitive on descriptive and action-oriented datasets, outperforming the 100K-scale Video-ChatGPT on MSVD-QA and MSRVTT-QA while maintaining strong performance on TGIF-QA, without sacrificing general video understanding.

\subsection{Ablation Analysis}

\begin{table}[t]
\centering
\caption{Ablation of V-CORE components on NExT-QA. ``Bi'' and ``Cau'' denote bidirectional and causal attention. ``Dyn. Sum.'' refers to Terminal Dynamic Summary. 
}
\label{tab:ablation_main}
\resizebox{\linewidth}{!}{
\begin{tabular}{c|cccc|c}
\toprule
\textbf{Config} & \textbf{Spatial} & \textbf{Temporal} & \textbf{Mask} & \textbf{Dyn. Sum.} & \textbf{Acc (\%)} \\
\midrule
(a) & MeanPool & Linear & - & - & 53.8 \\
(b) & LSA & Linear & - & - & 55.6 \\
(c) & LSA & Transformer & Bi & - & 58.0 \\
(d) & LSA & CATP & Cau & - & 60.2 \\
\rowcolor{gray!10} (e) & \textbf{LSA} & \textbf{CATP} & \textbf{Cau} & \checkmark & \textbf{61.2} \\
\bottomrule
\end{tabular}
}
\end{table}
\noindent\textbf{Progressive Component Ablation.}
Table~\ref{tab:ablation_main} quantifies the contribution of each component in V-CORE. 
Compared to the baseline (a), introducing LSA (b) and transformer-based temporal interaction (c) steadily improves performance by preserving salient spatial cues and enabling inter-frame modeling. 
Replacing bidirectional attention with our causal attention mask (d) yields a notable +2.2\% gain, directly validating our hypothesis that mitigating temporal leakage is critical for causal reasoning. 
Finally, incorporating the terminal Dynamic Summary (e) further boosts accuracy to 61.2\%, indicating that a learnable causal sink provides a more coherent global representation for long-range temporal reasoning.

\begin{table}[t]
\centering
\caption{
Detailed performance breakdown on NExT-QA sub-tasks for representative model variants.
}
\label{tab:subtask_breakdown}
\begin{tabular}{l|ccc|c}
\toprule
\textbf{Config} & \textbf{Tem.} & \textbf{Des.} & \textbf{Cau.} & \textbf{Avg.} \\
\midrule
(a) MeanPool + Linear & 49.2 & 62.1 & 50.1 & 53.8 \\
(c) LSA + Transf. (Bi) & 54.1 & 64.2 & 55.7 & 58.0 \\
\rowcolor{gray!10} (e) \textbf{V-CORE (Ours)} & \textbf{57.6} & \textbf{65.1} & \textbf{60.9} & \textbf{61.2} \\
\bottomrule
\end{tabular}
\end{table}
\noindent\textbf{Granular Performance Breakdown.}
Table~\ref{tab:subtask_breakdown} presents a sub-category analysis on NExT-QA.
Comparing the base (a), intermediate (c), and full (e) models, we observe that all configurations achieve comparable performance on descriptive questions, while V-CORE exhibits substantially larger gains on temporal (Tem.) and causal (Cau.) subsets. In particular, the full model improves causal accuracy by +5.2\% and temporal accuracy by +3.5\% over the bidirectional baseline (c), indicating that the performance gain primarily stems from enforcing causal temporal consistency rather than generic feature enhancement.

\begin{figure}[t]
    \centering
    \begin{subfigure}[b]{0.48\linewidth}
        \centering
        \includegraphics[width=\linewidth]{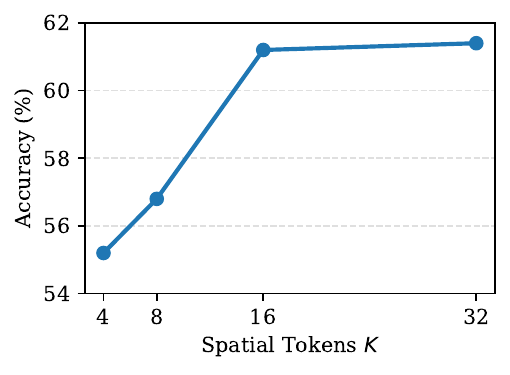}
        \caption{Impact of spatial tokens $K$}
        \label{fig:k_sensitivity}
    \end{subfigure}
    \hfill
    \begin{subfigure}[b]{0.48\linewidth}
        \centering
        \includegraphics[width=\linewidth]{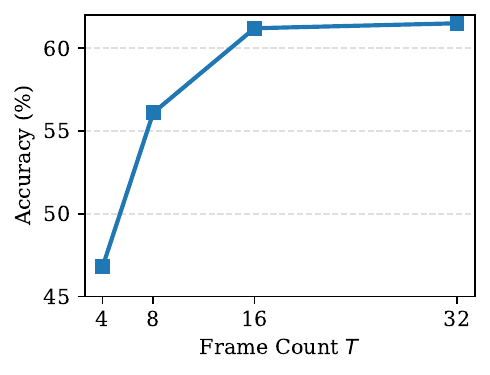}
        \caption{Impact of frame count $T$}
        \label{fig:t_sensitivity}
    \end{subfigure}
    \caption{Hyperparameter sensitivity on NExT-QA with respect to the spatial token count $K$ and the frame count $T$.}
    \label{fig:kt_sensitivity}
\end{figure}
\noindent\textbf{Hyperparameter Sensitivity.}
Fig.~\ref{fig:kt_sensitivity} studies the impact of the spatial token count $K$ and the frame count $T$ on NExT-QA. As shown in Fig.~\ref{fig:kt_sensitivity}(a), increasing $K$ improves accuracy up to a moderate value (e.g., $K=16$), after which performance saturates, indicating that a compact set of salient spatial tokens is sufficient. Fig.~\ref{fig:kt_sensitivity}(b) shows that accuracy increases rapidly as $T$ grows to 16 frames and plateaus thereafter. Overall, these results suggest that V-CORE achieves robust temporal reasoning with compact $(K, T)$ configurations.

\begin{table}[t]
\centering
\caption{
Ablation on the design of the summary token. ``Pre.'' and ``Ter.'' denote prepend and terminal positions.
}
\label{tab:ablation_summary_design}
\begin{tabular}{lllc}
\toprule
\textbf{Summary Design} & \textbf{Position} & \textbf{Aggregation} & \textbf{Acc (\%)} \\
\midrule
None (w/o Summary) & - & - & 60.2 \\
Static AvgPool & Ter. & AvgPool & 60.6 \\
Dynamic Query & Pre. & Learnable & 57.1 \\
\rowcolor{gray!10} \textbf{Dynamic Query (Ours)} & \textbf{Ter.} & \textbf{Learnable} & \textbf{61.2} \\
\bottomrule
\end{tabular}
\end{table}
\noindent\textbf{Analysis of Terminal Dynamic Summary Design.}
Table~\ref{tab:ablation_summary_design} validates the design of the terminal dynamic summary. 
Compared to static AvgPool, a learnable query at the terminal position yields a +0.6\% gain, indicating more effective global aggregation. In contrast, placing the query at the prepend position leads to a significant drop under causal masking, as it cannot access future frames. These results confirm that the summary token is most effective when acting as a causal sink at the end of the temporal reasoning chain.

\begin{figure*}[t]
    \centering
    \begin{subfigure}[t]{0.48\linewidth}
        \centering
        \includegraphics[width=\linewidth]{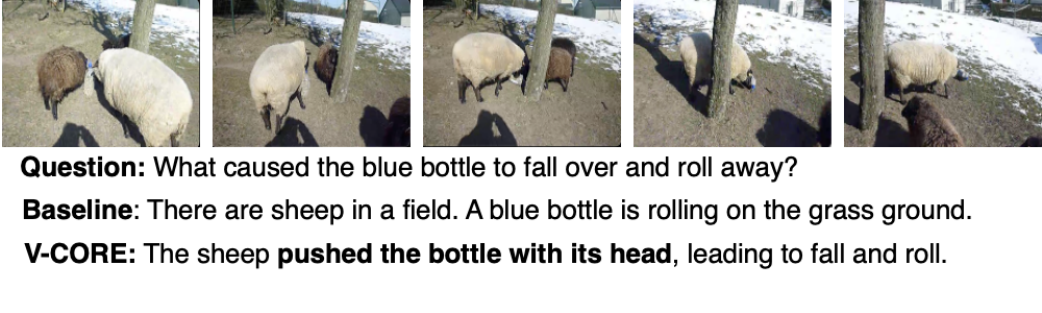} 
        \caption{\textbf{Case 1:} V-CORE identifies the head-butt as the trigger, while the baseline only describes the rolling motion.}
        \label{fig:case_causal}
    \end{subfigure}
    \hfill
    \begin{subfigure}[t]{0.48\linewidth}
        \centering
        \includegraphics[width=\linewidth]{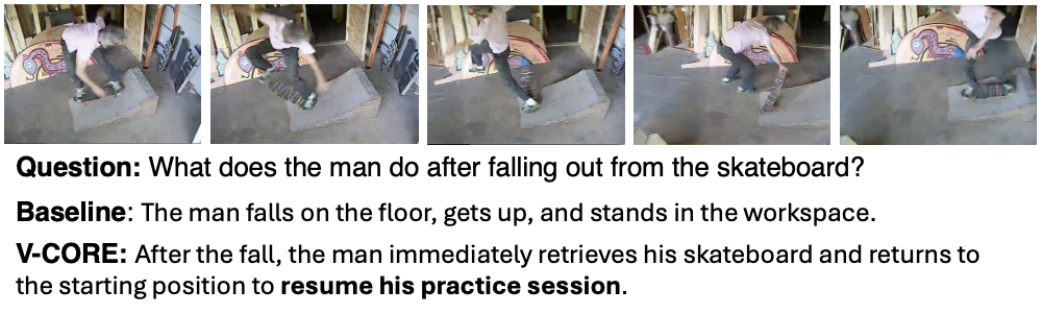} 
        \caption{\textbf{Case 2:} V-CORE captures goal-persistence by recognizing that the man resumes practice after a fall.}
        \label{fig:case_summary}
    \end{subfigure}
    
    \caption{Qualitative comparison between the baseline and \textbf{V-CORE}. Each case illustrates how our model leverages temporal constraints and global context to outperform baseline.}
    \label{fig:case_study}
\end{figure*}

\subsection{Qualitative Analysis}
Fig.~\ref{fig:case_study} presents qualitative comparisons between a baseline and \textbf{V-CORE}. 
In the first case, V-CORE correctly traces the causal antecedent of the event by grounding its reasoning in earlier frames, while the baseline focuses on post-event observations. 
In the second case, V-CORE produces a coherent global summary by aggregating long-range context, demonstrating improved narrative consistency under temporal constraints.


\section{Conclusion}
We presented V-CORE, a parameter-efficient framework for temporally consistent video understanding. By addressing temporal leakage in bidirectional projectors, V-CORE enforces explicit temporal flow through Learnable Spatial Aggregation and a Causality-Aware Temporal Projector with unidirectional attention. Under a rigorous zero-shot protocol, V-CORE achieves state-of-the-art performance on NExT-QA (61.2\% accuracy) while remaining efficient on a single consumer GPU. These results highlight the importance of explicit temporal constraints for robust and logically consistent video reasoning.

\section*{Limitations}
V-CORE has several limitations that suggest directions for future work. First, the use of segment-based uniform sampling may overlook fine-grained events in extremely long or high-tempo videos; adaptive frame selection could improve coverage. Second, the fixed number of compressed tokens ($TK+1$) reflects a trade-off between efficiency and representational granularity for dense narratives. Finally, our evaluation focuses on Video Question Answering; extending V-CORE to open-ended video dialogue remains future work.

\bibliographystyle{IEEEbib}

\begingroup
\sloppy
\setlength{\emergencystretch}{2em}
\bibliography{icme2026references}

@inproceedings{maaz2024videochatgpt,
  title={Video-chatgpt: Towards detailed video understanding via large vision and language models},
  author={Maaz, Muhammad and Rasheed, Hanoona and Khan, Salman and Khan, Fahad},
  booktitle={ACL},
  pages={12585--12602},
  year={2024}
}

@article{zhang2023videollama,
  title={Video-llama: An instruction-tuned audio-visual language model for video understanding},
  author={Zhang, Hang and Li, Xin and Bing, Lidong},
  journal={arXiv preprint arXiv:2306.02858},
  year={2023}
}

@inproceedings{lin2023videollava,
  title={Video-llava: Learning united visual representation by alignment before projection},
  author={Lin, Bin and Ye, Yang and Zhu, Bin and Cui, Jiaxi and Ning, Munan and Jin, Peng and Yuan, Li},
  booktitle={EMNLP},
  pages={5971--5984},
  year={2024}
}

@article{li2023videochat,
  title={Videochat: Chat-centric video understanding},
  author={Li, KunChang and He, Yinan and Wang, Yi and Li, Yizhuo and Wang, Wenhai and Luo, Ping and Wang, Yali and Wang, Limin and Qiao, Yu},
  journal={arXiv preprint arXiv:2305.06355},
  year={2023}
}

@article{wang2024qwen2vl,
  title={Qwen2-vl: Enhancing vision-language model's perception of the world at any resolution},
  author={Wang, Peng and Bai, Shuai and Tan, Sinan and Wang, Shijie and Fan, Zhihao and Bai, Jinze and Chen, Keqin and Liu, Xuejing and Wang, Jialin and Ge, Wenbin and others},
  journal={arXiv preprint arXiv:2409.12191},
  year={2024}
}

@inproceedings{bertasius2021timesformer,
  title={Is space-time attention all you need for video understanding?},
  author={Bertasius, Gedas and Wang, Heng and Torresani, Lorenzo},
  booktitle={ICML},
  volume={2},
  number={3},
  pages={4},
  year={2021}
}

@inproceedings{arnab2021vivit,
  title={Vivit: A video vision transformer},
  author={Arnab, Anurag and Dehghani, Mostafa and Heigold, Georg and Sun, Chen and Lu{\v{c}}i{\'c}, Mario and Schmid, Cordelia},
  booktitle={CVPR},
  pages={6836--6846},
  year={2021}
}

@inproceedings{li2024llamavid,
  title={Llama-vid: An image is worth 2 tokens in large language models},
  author={Li, Yanwei and Wang, Chengyao and Jia, Jiaya},
  booktitle={ECCV},
  pages={323--340},
  year={2024},
  organization={Springer}
}

@inproceedings{hu2021lora,
  title={LoRA: Low-Rank Adaptation of Large Language Models},
  author={Hu, Edward J and Shen, Yelong and Wallis, Phillip and Allen-Zhu, Zeyuan and Li, Yuanzhi and Wang, Shean and Wang, Lu and Chen, Weizhu},
  booktitle={ICLR},
  year={2022}
}

@article{dettmers2023qlora,
  title={QLoRA: Efficient Finetuning of Quantized LLMs},
  author={Dettmers, Tim and Pagnoni, Artidoro and Holtzman, Ari and Zettlemoyer, Luke},
  journal={arXiv preprint arXiv:2305.14314},
  year={2023}
}

@article{achiam2023gpt4,
  title={Gpt-4 technical report},
  author={Achiam, Josh and Adler, Steven and Agarwal, Sandhini and Ahmad, Lama and Akkaya, Ilge and Aleman, Florencia Leoni and Almeida, Diogo and Altenschmidt, Janko and Altman, Sam and Anadkat, Shyamal and others},
  journal={arXiv preprint arXiv:2303.08774},
  year={2023}
}

@article{touvron2023llama2,
  title={Llama 2: Open Foundation and Fine-Tuned Chat Models},
  author={Touvron, Hugo and Martin, Louis and Stone, Kevin and others},
  journal={arXiv preprint arXiv:2307.09288},
  year={2023}
}

@article{liu2023llava,
  title={Visual instruction tuning},
  author={Liu, Haotian and Li, Chunyuan and Wu, Qingyang and Lee, Yong Jae},
  journal={NIPS},
  volume={36},
  pages={34892--34916},
  year={2023}
}

@inproceedings{li2023blip2,
  title={Blip-2: Bootstrapping language-image pre-training with frozen image encoders and large language models},
  author={Li, Junnan and Li, Dongxu and Savarese, Silvio and Hoi, Steven},
  booktitle={ICML},
  pages={19730--19742},
  year={2023},
  organization={PMLR}
}

@inproceedings{jang2017tgif,
  title={Tgif-qa: Toward spatio-temporal reasoning in visual question answering},
  author={Jang, Yunseok and Song, Yale and Yu, Youngjae and Kim, Youngjin and Kim, Gunhee},
  booktitle={CVPR},
  pages={2758--2766},
  year={2017}
}

@inproceedings{xiao2021next,
  title={Next-qa: Next phase of question-answering to explaining temporal actions},
  author={Xiao, Junbin and Shang, Xindi and Yao, Angela and Chua, Tat-Seng},
  booktitle={CVPR},
  pages={9777--9786},
  year={2021}
}

@inproceedings{xu2016msr,
  title={Msr-vtt: A large video description dataset for bridging video and language},
  author={Xu, Jun and Mei, Tao and Yao, Ting and Rui, Yong},
  booktitle={CVPR},
  pages={5288--5296},
  year={2016}
}

@inproceedings{ye2017video,
  title={Video question answering via attribute-augmented attention network learning},
  author={Ye, Yunan and Zhao, Zhou and Li, Yimeng and Chen, Long and Xiao, Jun and Zhuang, Yueting},
  booktitle={ACM SIGIR},
  pages={829--832},
  year={2017}
}

@article{wang2022internvideo,
  title={Internvideo: General video foundation models via generative and discriminative learning},
  author={Wang, Yi and Li, Kunchang and Li, Yizhuo and He, Yinan and Huang, Bingkun and Zhao, Zhiyu and Zhang, Hongjie and Xu, Jilan and Liu, Yi and Wang, Zun and others},
  journal={arXiv preprint arXiv:2212.03191},
  year={2022}
}

@article{zhang2024llavanext,
  title={LLaVA-NeXT: Stronger LLMs Ignite Multimodal Video Learning},
  author={Zhang, Yuanhan and Li, Bo and Liu, Haotian and et al.},
  journal={arXiv preprint arXiv:2404.16821},
  year={2024}
}

@inproceedings{li2024mvbench,
  title={Mvbench: A comprehensive multi-modal video understanding benchmark},
  author={Li, Kunchang and Wang, Yali and He, Yinan and Li, Yizhuo and Wang, Yi and Liu, Yi and Wang, Zun and Xu, Jilan and Chen, Guo and Luo, Ping and others},
  booktitle={CVPR},
  pages={22195--22206},
  year={2024}
}

@inproceedings{liu2024stllm,
  title={St-llm: Large language models are effective temporal learners},
  author={Liu, Ruyang and Li, Chen and Tang, Haoran and Ge, Yixiao and Shan, Ying and Li, Ge},
  booktitle={ECCV},
  pages={1--18},
  year={2024},
  organization={Springer}
}

@inproceedings{fu2024videomme,
  title={Video-mme: The first-ever comprehensive evaluation benchmark of multi-modal llms in video analysis},
  author={Fu, Chaoyou and Dai, Yuhan and Luo, Yongdong and Li, Lei and Ren, Shuhuai and Zhang, Renrui and Wang, Zihan and Zhou, Chenyu and Shen, Yunhang and Zhang, Mengdan and others},
  booktitle={CVPR},
  pages={24108--24118},
  year={2025}
}

@inproceedings{ma2024vista,
  title={Vista-llama: Reducing hallucination in video language models via equal distance to visual tokens},
  author={Ma, Fan and Jin, Xiaojie and Wang, Heng and Xian, Yuchen and Feng, Jiashi and Yang, Yi},
  booktitle={CVPR},
  pages={13151--13160},
  year={2024}
}

@inproceedings{deng2024covis,
  title={Covis: A collaborative framework for fine-grained graphic visual understanding},
  author={Deng, Xiaoyu and Kang, Zhengjian and Li, Xintao and Zhang, Yongzhe and Guo, Tianmin},
  booktitle={2025 28th International Conference on Computer Supported Cooperative Work in Design (CSCWD)},
  pages={298--303},
  year={2025},
  organization={IEEE}
}

@inproceedings{kang2025lpdetr,
  title={Lp-detr: Layer-wise progressive relation for object detection},
  author={Kang, Zhengjian and Zhang, Ye and Deng, Xiaoyu and Li, Xintao and Zhang, Yongzhe},
  booktitle={International Conference on Intelligent Computing},
  pages={144--156},
  year={2025},
  organization={Springer}
}

@inproceedings{yu2024spatial,
  title={Spatial Transform Decoupling for Oriented Object Detection},
  author={Yu, Hongtian and Tian, Yunjie and Ye, Qixiang and Liu, Yunfan},
  booktitle={Proceedings of the AAAI Conference on Artificial Intelligence},
  volume={38},
  number={7},
  pages={6782--6790},
  year={2024}
}

@inproceedings{li2025mcl,
  title={Mcl for mllms: Benchmarking forgetting in task-incremental multimodal learning},
  author={Li, Zichao},
  booktitle={ICCV Workshop},
  pages={2760--2766},
  year={2025}
}

@inproceedings{cui2023robustnesslargemultimodalmodels,
  title={On the robustness of large multimodal models against image adversarial attacks},
  author={Cui, Xuanming and Aparcedo, Alejandro and Jang, Young Kyun and Lim, Ser-Nam},
  booktitle={CVPR},
  pages={24625--24634},
  year={2024}
}

@article{kang2026paq,
  title={PaQ-DETR: Learning Pattern and Quality-Aware Dynamic Queries for Object Detection},
  author={Kang, Zhengjian and Zhuang, Jun and Mo, Kangtong and Chen, Qi and Liu, Rui and Zhang, Ye},
  journal={arXiv preprint arXiv:2603.06917},
  year={2026}
}

@InProceedings{Li_2026_WACV,
    author    = {Li, Zichao},
    title     = {Probabilistic Scene Graph Prompting: Uncertainty-Aware Structured Reasoning in Multimodal LLMs},
    booktitle = {Proceedings of the IEEE/CVF Winter Conference on Applications of Computer Vision (WACV) Workshops},
    year      = {2026},
    pages     = {1695-1702}
}

@inproceedings{xing2025openemma,
  title={Openemma: Open-source multimodal model for end-to-end autonomous driving},
  author={Xing, Shuo and Qian, Chengyuan and Wang, Yuping and Hua, Hongyuan and Tian, Kexin and Zhou, Yang and Tu, Zhengzhong},
  booktitle={WACV},
  pages={1001--1009},
  year={2025}
}
\endgroup


\end{document}